\documentclass[runningheads]{llncs}

\usepackage{url}
\usepackage[hidelinks]{hyperref}
\usepackage[utf8]{inputenc}
\usepackage[small]{caption}
\usepackage{graphicx}
\usepackage{amsmath}
\usepackage{booktabs}
\usepackage{algorithm}
\usepackage{algorithmic}
\usepackage{pdfpages}
\usepackage{amssymb}
\usepackage{multirow}

\newcommand{\name}{VIKING}
\newcommand*\rot{\rotatebox{90}}

\DeclareMathOperator*{\argmax}{argmax}
\DeclareMathOperator*{\argmin}{argmin}

\begin{document}

\title{VIKING: Adversarial Attack on Network Embeddings via Supervised Network Poisoning}
\titlerunning{VIKING}

\author{Viresh Gupta \and Tanmoy Chakraborty}
\institute{
    Indraprastha Institute of Information Technology Delhi, India
    \email{\{viresh16118,tanmoy\}@iiitd.ac.in}
 }

\maketitle

\begin{abstract}
Learning low-level node embeddings using techniques from network representation learning is useful for solving downstream tasks such as node classification and link prediction. An important consideration in such applications is the robustness of the embedding algorithms against adversarial attacks, which  can be examined by performing perturbation on the original network. An efficient perturbation technique can degrade the performance of network embeddings on  downstream tasks. In this paper, we study network embedding algorithms from an adversarial point of view and observe the effect of poisoning the network on downstream tasks. We propose \name, a supervised network poisoning strategy that outperforms the state-of-the-art poisoning methods by up to $18\%$ on the original network structure. We also extend \name\ to a semi-supervised attack setting and show that it is comparable to its supervised counterpart.

\end{abstract}
\section{Introduction}

Several network analysis problems involve prediction over nodes and edges in a network. 

Traditional network science methods achieve this by analyzing networks using graph properties (such as centrality based techniques) or network factorization \cite{chen2018tutorial}. These methods are slow and do not scale very well. Advancements in NLP, especially word vectors \cite{word2vec} have inspired several embedding algorithms for graph data as well. This family of embedding algorithms uses random walks to sample neighboring nodes and optimizes node embeddings to minimize a predefined objective (e.g. Deepwalk \cite{deepwalk} and Node2Vec \cite{node2vec}). 

Even though network embeddings have recently gained considerable attention, embedding algorithms have not been studied to check their robustness to attacks in the network. Network attacks can generally be classified into {\em whitebox attacks} (where all information including parameters of the learned classifier and the model used are known) and {\em blackbox attacks} (where no information about the system is known; however, access to predictions is available). Blackbox attacks on networks can happen in two ways, viz. poisoning and evasion attacks. The poisoning attacks modify the network before the algorithm/model is trained; whereas evasion attacks focus on fooling a model after it has been trained \cite{chakraborty2018adversarial}.

In this work, we focus on developing a black-box attack method (which is blind to the embedding algorithm used) and achieve it via network poisoning. Although adversarial attacks can happen in several ways, as far as undirected, unattributed, and unweighted networks are concerned, the only applicable attack is either the addition of an edge or the removal of an edge. These attacks are simply referred to as {\bf edge flipping}. A significant portion of studies pertains to developing adversarial attacks on image data.
However, several recent efforts \cite{zgner2019adversarial,ZugnerKDD,dai2018adversarial} have also shed light on adversarial attacks on semi-supervised network learning models such as Graph Convolutional Networks (GCNs) \cite{gcn}. However, there has been a lack of studies that explore the effect of perturbations on an unsupervised node embedding method. We are particularly interested in probing how the absence of a supervision signal during embedding affects such attacks. 

We focus on homogeneous non-attributed networks, i.e., networks in which nodes do not have any attribute; all nodes represent the same kind of entity; and all edges are undirected and unweighted. We propose adversarial attack on network embeddings via {\bf super\underline{vi}sed networ\underline{k} poison\underline{ing} (\name)}, which, unlike previous strategies, incorporates supervision during attack time \textcolor{black}{(in the form of node labels)}. Our investigation shows that \name, even in a semi-supervised setting, is extremely effective. It is able to degrade the performance of the unsupervised embedding methods (and supervised embedding methods like GCN as well), leading to a decrease in the micro F1-score for node classification (up to $19 \%$ when applied on synthetic networks and up to $18 \%$ when applied on three real-world networks). \name\ also performs efficiently on the link prediction task, decreasing average precision by up to $50 \%$. The semi-supervised counterpart of \name\ performs similar to the other baseline methods, demonstrating the possibility of attacks even with partial knowledge.

In short, our contributions are summarized below:
\begin{itemize}
    \item We develop \name\, a generic adversarial attacking framework for discrete network features.
    \item We quantify the assumption of homophily behind random walk based embedding methods using node labels.
    \item We develop a supervised attacking strategy using the above label-based heuristic.
    \item We extend \name\ to a semi-supervised setting (VIKING$^s$) and show that it is equally effective.
\end{itemize}

{\bf Reproducibility:} The code and the datasets are public at the following link: \url{https://github.com/virresh/viking}.

\section{Related work \label{relatedwork}}
We focus on scalable node embedding approaches (mostly based on random walks) due to their flexibility for downstream tasks. For a survey on embedding methods and the advancements, readers are referred to \textcolor{black}{\cite{GOYAL201878_survey}}. In the present work, our focus is on analyzing the vulnerability to adversarial attacks of those methods that do not have access to supervision signals for embedding (like Deepwalk \cite{deepwalk}, Node2Vec \cite{node2vec}, LINE \cite{node2vec}) or partial access (like GCN \cite{gcn}). 

Initial adversarial poisoning attacks were against Support Vector Machines (SVMs) and traditional machine learning models \cite{Meipoisoning,biggio2012poisoning}. Many studies showed neural networks to be vulnerable to slight perturbations in the input data \cite{goodfellow2014explaining}. Although a significant portion of such studies focused on image data \cite{deepfool}, recent efforts have also been directed towards other kinds of data \cite{Liang_2018_text_fool,audio_fool}. 
Attacks on networks have often focused on exploiting some model parameters. In \cite{ZugnerKDD} GCN is linearized for attacking, \cite{dai2018adversarial} creates adversarial examples using reinforcement learning approach, \cite{zgner2019adversarial} exploits meta-gradients for attacking. However, \cite{zugnerPoisoning} develops a poisoning strategy for attacking networks, they only leverage the information of loss function. It is possible to use supervision to degrade the performance of embedding methods; however to the best of our knowledge, no study has investigated such supervised attack methods for random walk based embedding approaches.
\section{Preliminaries \label{attackmodel}}
In order to design an adversarial attack, we take the formal problem definition from \cite{zugnerPoisoning}, which is a bi-level optimization problem and break it down to a relatively easier problem. To limit the attacker's activities, we  assume a budget restriction. We focus on \textcolor{black}{a variety of embedding approaches such as -- random walk based approaches (DeepWalk and Node2Vec), large scale embedding algorithm (LINE) and a semi-supervised embedding method (GCN)}. We consider networks denoted by $G = (V,E)$ such that they are unweighted, unattributed and undirected, with $V$ denoting the set of vertices and $E$ denoting the set of edges. \textcolor{black}{In order to focus on attacking the graph's structural information, we only focus on poisoning the original network connections by flipping edges.} 

Let $A \in \{0, 1\}^{|V|\times|V|}$ be the adjacency matrix associated with network $G$. The goal of learning network embeddings is to learn a representation $Z = [z_v] \in R^{K}$ for each node $v\in V$ such that $K<<|V|$. Such embeddings $(Z)$ are then used by a downstream function $\zeta$ to perform an end task. 
As an attacker, our goal is to flip some values in $A$ so that the new adjacency matrix $A'$ results in learning embeddings $Z' = [z'_{v}]$ which in turn results in comparatively worse performance for the downstream function $\zeta$.

In most practical scenarios, there are constraints to the amount of perturbation allowed on the network. Therefore,  we fix a budget of $b$ flips in total. Every flip $f$ changes $A$, such that $||A' - A||_0 = 2$. Thus, this problem boils down to the following bi-level optimization:

\begin{equation}\label{eqn:bilevelA}
(A')^{\ast} = \argmax_{A'} {\Delta}(A', Z^{\ast})
\end{equation}
\begin{equation}\label{eqn:bilevelB}
Z^{\ast} = \argmin_{Z} {\zeta}(A', Z)
\end{equation}
subject to $|A'-A|_0 = 2b$. ${\Delta}$ is the loss function that we will have to design specific to the problem and ${\zeta}$ is the embedding \textcolor{black}{algorithm's loss}.

Solving this optimization problem is challenging. We shall solve this bi-level optimization problem approximately by converting it to a single optimization problem and use a simple brute-force solution. \textcolor{black}{To achieve this, we will decouple the embeddings $Z^*$ from the attacking function and replace it with a proxy (defined in Section \ref{genericmodel}).}

\section{Network poisoning strategy}
In this section, we begin by describing the general attacking framework to decide which edges to flip, \textcolor{black}{followed by our approach where we plug in a parameter in the general attacking framework that makes use of supervision in the time of attack, and then propose a simple semi-supervised extension.}
\subsection{Generic poisoning framework \label{genericmodel}}
Adding or removing an edge leads to the flipping of two symmetric entries in an adjacency matrix. We form a candidate edge set from which some edges will be chosen for addition or removal as required. The procedure to compute the candidate edge set is as follows:

\begin{itemize}
    \item {\bf Edges to be removed:} We randomly mark one edge attached with each node as \textit{safe}. These edges will not be removed during the poisoning attack. All the {\em unsafe} edges now belong to our candidate edge set. This ensures that there are no isolated vertices introduced in the resultant network even if we remove all the edges from the candidate edge set. \textcolor{black}{However, if there were isolated vertices present in the network, then they will remain as is}.
    \item {\bf Edges to be added:} Since all non-edges in the original network can be potentially added, we compute the adjacency matrix of the network complement and then include each edge in the upper triangular portion of the matrix (to avoid repetitions) in our potential edge addition set. Since this set is generally large, we randomly sample a reasonable number of candidate edges from this set.
\end{itemize}

As mentioned before, we need to constrain our attack with a budget of $b$ edge flips because it is impractical for an attacker to remove all the edges from a network or add all missing edges to a network. Accordingly, we assign an importance value $imp_e$ to every edge of the candidate set. \textcolor{black}{This importance value shall determine the top $b$ edges to be flipped and give us the optimized poisoned network $A'$ (i.e., the adjacency matrix)}. 

\textcolor{black}{In the general attack strategy, there can be several ways to determine $imp_e$; however, we discuss here a simple method to compute the \textit{importance value} for each candidate edge with respect to a particular graph property/feature $d$}. Suppose there is a node feature $d$ that we want to attack (e.g., node centrality or degree). Let us restrict $d$ to be a discrete variable having at most $D$ possible values  (in case of continuous values, we can just set this to the dimensionality of the feature). We can then represent a feature matrix as $F \in \{0, 1\}^{|V|\times D}$. The key insight into developing this attack is to observe that embeddings aim to preserve similarity in the embedding space. Thus, if all nodes have the same feature value, then that feature becomes useless in the embedding space.

Hence, our objective is to make this feature value of each node as close to each other as possible. For this purpose, we compute $imp_e$ of each candidate edge flip by some function (let's say $\theta$) of the feature/property $d$ with respect to the new attacked graph $A'$. i.e:
\begin{equation} \label{eqn:generic}
    imp_e = C - \theta(F, A')
\end{equation} 
where $C$ is a constant and $\theta$ is our defined loss.

\textcolor{black}{In order to compute $\theta$, we can ``aggregate" the neighbour features by multiplying the adjacency matrix with feature matrix and weight them elementwise by the original node's feature value ($(A\times F) \circ F$); \textcolor{black}{where $\circ$ represents element-wise multiplication}. Since this vector is arbitrary, we will normalize this with the unweighted equivalent ($A\times F$) and compare the distance (l2 norm) with the normalized importance of the original graph, in order to call it $imp_e$. Note that even though this $imp_e$ is for determining the importance of a candidate edge, it assigns a value to the whole graph after the edge $e$ was flipped.}  These operations can be summarized as follows
\newcommand{\diag}{\mathop{\mathrm{diag}}}
\begin{equation}
   \theta(F, A') = ||(\diag(\sum (A' \times F)))^{-1}\sum ((A' \times F) \circ F)||_2
\end{equation}
where $\sum$ represents the summation across first dimension (resulting in a vector of $|V| \times 1$ for $G = (V, E)$). $A'$ is the poisoned graph after flipping exactly one edge $e$ and $F$ is the feature matrix as defined above. Since we are comparing the importance across several flips with the original graph ($A$), we treat $C$ as the importance of the original graph, i.e:
    $C = \theta(F, A)$.

We describe the motivation for choosing such a loss function. The multiplication of two binary matrices $A \in \{0, 1\}^{|V|\times |V|}$ and $F \in \{0, 1\}^{|V|\times D}$ produces a new matrix $AF \in \mathbb{W}^{ |V| \times D}$ with features accumulated from its neighbors. Element-wise multiplication with $F$ again weights out the aggregated values for each node by its own feature value. A row-wise summation $\sum (AF \circ F) \in \mathbb{W}^{|V|}$ provides a numeric characterization with which we can approximate its closeness to a given value. The division by $\sum AF$ is done for normalization and allows computing distance of the numeric characterization using a constant value.

\textbf{Importance of using constant} ${C}$: We aim to bring the feature values of each node as close to each other as possible. One way of achieving this would be to bring the numeric characterization of feature values for each node as close to a fixed constant as possible. This will automatically imply bringing the values for each node closer to each other. Thus ${C}$ is a \textcolor{black}{scalar} which allows us to choose the value at which we want our feature values in $F$ to converge. \textcolor{black}{This is how we propose to achieve the objective of ``making node embeddings less useful by making them similar"}.

Now we can convert our bi-level optimisation to a single objective -- minimizing $imp_e$. This is the proxy that can be used instead of embeddings $Z^*$ in equation \ref{eqn:bilevelA}. As for solving the objective $\Delta$, we can simply use the following brute force approach:\\
For every edge flip $e$, we can assign an importance value to the attacked graph ($A'$) with $imp_e$ and then sort the candidate edge set in descending order of the importance values. Now we can pick top $b$ edges from this sorted candidate edge set to obtain the final attacked graph $(A')^*$.

\textcolor{black}{To summarize, the higher the importance of a flip, the more number of node features will be brought closer on flipping that edge}.

This method constitutes our generic attack framework. We can use a known node importance measure instead assuming a generic feature $d$, such as vertex degree or node centrality. \textcolor{black}{This framework can thus be used for both unsupervised, semi-supervised or supervised attacks. To use it in unsupervised fashion, one can use a node-property that can be computed from the graph. To use it in supervised or semi-supervised fashion, we just need to use an external information such as node-labels. We will show its utility as a supervised and semi-supervised method in the following sections.}

\subsection{VIKING: Our proposed poisoning strategy \label{supervisedpoisoning}}
In this section, we provide a logical choice for the feature matrix $F$. For this purpose, we first recall why the word vector approach works. As stated in \cite{word2vec}, embeddings try to capture similarity i.e., -- ``birds of a feather flock together". This is an instance of homophily which is exhibited in several real-world networks, and this is the insight that makes network embeddings a useful tool. Thus an ideal candidate for $F$ would be a parameter that quantifies homophily at every node.

For this purpose, let us define for every node an intracommunity-intercommunity ratio $\eta = \frac{n_{same}(x)}{n_{diff}(x)}$, where $n_{diff}(x)$ is the number of neighbors of $x$ that do not belong to the same community as $x$, and $n_{same}(x)$ is the number of neighbors of $x$ that belong to the same community as $x$. As shown in Section \ref{genericmodel}, the generic loss function makes $\eta$ converge to a constant value of our choice. 

Algorithm \ref{alg:saanp} summarizes VIKING. 
\textcolor{black}{
In order to estimate $\eta$, we only need to replace $F$ in Equation \ref{eqn:generic} by the community label matrix $F_{\eta}$ of the given graph (i.e., a one-hot representation of the community labels of every node in the graph), where $F_{\eta} \in \mathbb{W}^{|V|\times \alpha}$, $\alpha$ is the number of unique labels/communities, and $\mathbb{W}$ represents the set of whole numbers. This particular computation of $\eta$ is natively supported by our generic attack framework by simply plugging in the labels. This will result in the following objective:
\begin{equation}
\label{eqn:viking}
    (A')^* = \argmin_{A'} {C} - \theta(F_{\eta}, A')
\end{equation}
}

\begin{algorithm}[!t]
    \caption{VIKING algorithm}
    \label{alg:saanp}
    \textbf{Input}: Adjacency matrix $A$ of network, candidate set $CS$ for possible edge flips\\
    \textbf{Parameter}: Constant ${C}$, flip budget $b$ \\
    \textbf{Output}: Adjacency matrix $A'$ of the poisoned network
    \begin{algorithmic}[1]\small 
        \FORALL{flip $f$ in $CS$}
            \STATE Compute $imp_{f} = \theta(F_{\eta}, A')$
        \ENDFOR
        \STATE \textbf{sort} $CS$ on the basis of $\Delta_f$ in decreasing order
        \STATE \textbf{choose} $\mathrm{top}_f =$ First $b$ flips from sorted $CS$
        \STATE \textbf{apply} $\mathrm{top}_f$ to $A$
        \STATE \textbf{return} $A'$
    \end{algorithmic}
\end{algorithm}

{{\bf Time complexity:}} We choose to select $k|E|$ edges for addition (for some constant $k$) and at most $|E|$ edges for removal. Assuming that the average degree in the network is $\overline{d}$, the time complexity of VIKING is $O(|E||V|\overline{d})$ which is linear in the number of edges.

\subsection{VIKING$^s$: Semi-supervised extension}\label{vikingminus}
To extend VIKING into a semi-supervised setting, we apply a learned feature matrix $F_{\eta}'$ instead of using the ground-truth $F_{\eta}$. This can be done using the node-classification task. First, we generate initial unsupervised embeddings $Z_0$ using an embedding method (skipgram DeepWalk in this case), and use these embeddings in conjunction with a logistic regression classifier trained with $x\%$ of ground-truth community labels (we take $x=10$). The logistic regression is then used to predict $F_{\eta}'$, the surrogate label matrix to be used in place of $L$ in Equation \ref{eqn:viking}. The edges selected by VIKING$^s$ can then be flipped and the resultant poisoned graph can be used for different downstream tasks (node classification and link prediction).

\section{Datasets \label{datasets}}

We use two kinds of networks for the evaluation.
Table \ref{tab:datasets} provides the statistics of the networks.

\noindent{{\bf \underline{Synthetic networks}:}} We use two synthetic network generators:

(i) \textbf{LFR networks} are commonly used as benchmark  for community detection in networks \cite{lancichinetti2008benchmark}. We treat every community as a class/label to train a node classifier. 
We perform experiments on different networks by choosing the power law exponent for the degree distribution to be $3$, the power law exponent for the community size distribution to be $2$, the desired average degree of nodes to be $20$, the minimum size of communities to be $200$, and varying the fraction of intra-community edges incident to each node ($\mu$). 

\begin{table}[!t]
\footnotesize
\caption{Network statistics. The flip budget shown here is default.\label{tab:datasets}}
\centering
\begin{tabular}{@{}c|c|c|c|c@{}}
\toprule
Network          & \#nodes       & \#edges       & Flip budget (\%)   & \#communities \\   \midrule
LFR        & 1000        & 11000       & 1000 (9.09)         & 3            \\ 
Cora       & 2810        & 7981        & 1000 (12.25)         & 7           \\
ForestFire & 1000        & 2721        & 500 (18.37)          & 30          \\
PolBlogs   & 1222        & 16717       & 1000 (5.98)         & 2            \\
CiteSeer   & 2110        & 7388        & 1000 (18.74)         & 6           \\\bottomrule
\end{tabular}
\end{table}

\textcolor{black}{(ii) \textbf{Forest Fire} is another method for generating graphs. Since it doesn't directly provide us community structure, we use Louvain algorithm \cite{Blondel_2008} for creating community partitions. The graph is generated using the forest fire model \cite{leskovec2007graph}, for $1000$ nodes with parameters \textit{forward burning rate} of $0.4$ and a \textit{backward burning rate} of $0.2$.  
}

\noindent\textbf{\underline{Real-world networks:}}
We used three standard, publicly available datasets for this purpose. These datasets are the same as the ones used in \cite{zugnerPoisoning}. The first of them is the Cora dataset which is a citation network. The second dataset is PolBlogs. It is a network of political blogs belonging to either rightwing or leftwing, leading to binary classification. The third dataset is the CiteSeer dataset which is also a citation network. 

\section{Experimental evaluation \label{experimentalevaluation}}

The evaluation of our poisoning strategy is done using two downstream tasks -- node classification and link prediction. For node-classification, network embeddings of the poisoned network are generated using an unsupervised embedding method and used as features within a simple logistic regression. The logistic regression is trained on the network embeddings obtained after poisoning the network. Only 10\% of the available node labels are used for training logistic regression. Micro F1-scores are reported by averaging over 10 runs. Similarly for link-prediction, we compute cosine similarity of embeddings for node pairs and use it as prediction score for computing average precision (AP).

For evaluation, we use two techniques of generating candidate edges discussed in Section \ref{attackmodel}, viz addition and removal. Additionally, we combine both these in a combined strategy to observe the effect of using both the candidates together in the experiments.

\noindent\textbf{\underline{Competing methods}:} The following are the competing methods: (i) ``Clean" refers to the clean network before poisoning; (ii) ``Random" refers to the random edge flip baseline; (iii) ``UNSUP" refers to the unsupervised poisoning baseline used in \cite{zugnerPoisoning}; (iv) ``\name" refers to results when the network is poisoned using our proposed method (Section \ref{supervisedpoisoning}); and (v) ``\name$^s$" refers to the semi-supervised version of \name\ (Section \ref{vikingminus}).

\noindent\textbf{\underline{Embedding methods}:} We run all experiments using three \textcolor{black}{unsupervised} embedding algorithms --  DeepWalk, Node2Vec, and LINE. For DeepWalk, we use two variants -- the Skipgram version \cite{deepwalk} and the SVD version \cite{Qiu:2018:NEM:3159652.3159706}. The SVD version of DeepWalk approximates the objective of random walks by using matrix factorization. 
For Node2Vec, we use the author's original implementation provided in the SNAP package (\url{snap.stanford.edu/node2vec/}). Similarly for LINE, we use the author's original implementation(\url{github.com/tangjianpku/LINE}). 

\textcolor{black}{We also use a semi-supervised method -- Graph Convolutional Network (GCN, \cite{gcn}). For GCN, we use a two level network with the middle layer's size equal to the embedding dimension and final layer's size equal to the number of communities. These are trained with the node labels; hence these embeddings do not perform well on the link-prediction task relative to the node community labels. } \textcolor{black}{The GCN network is implemented with the help of DGL library \cite{wang2019dgl}.}

\begin{figure}[!t]
    \centering
    \includegraphics[width=0.8\textwidth]{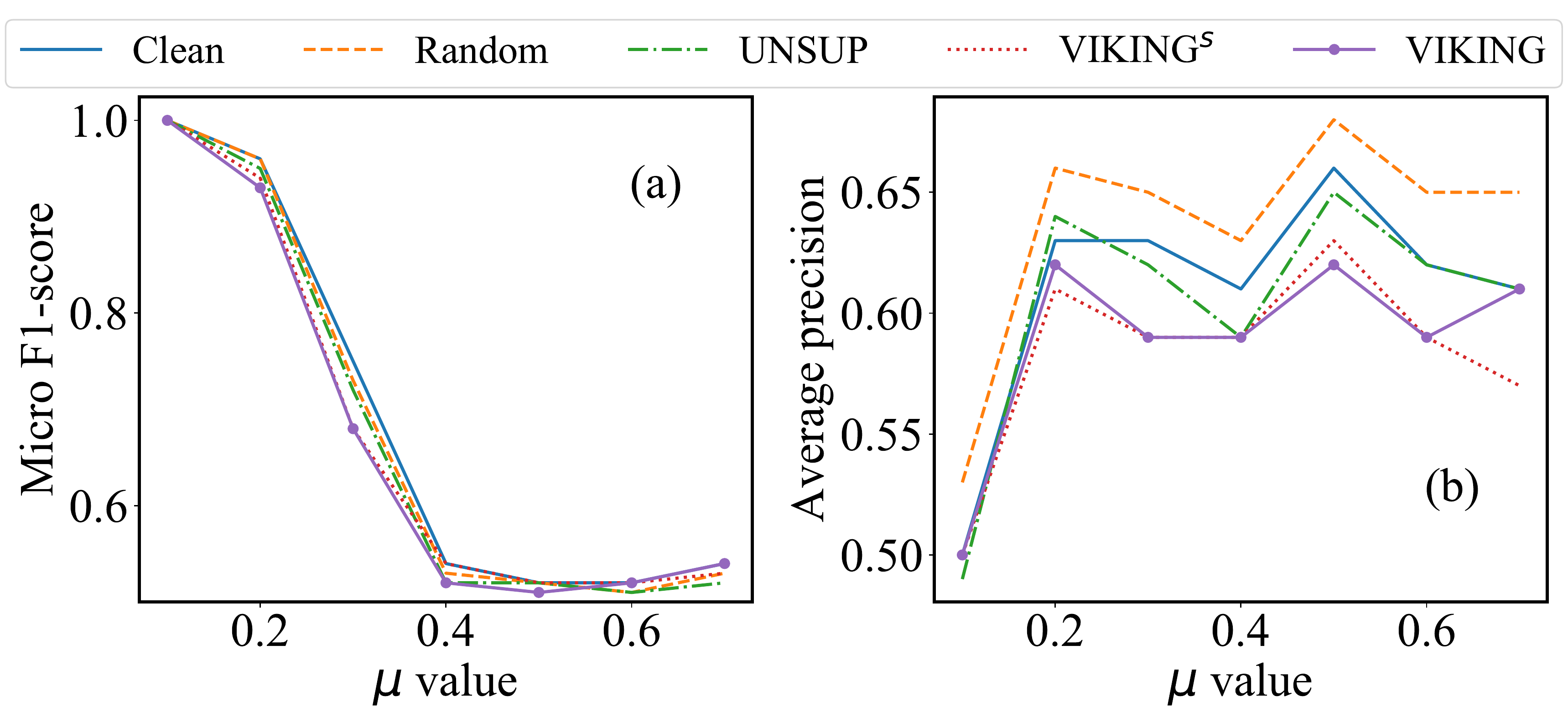}
      \caption{Variation of performance with $\mu$ on LFR networks for (a) node classification, and (b) link prediction. $\mu$ varies from 0.1 to 0.7. Note that on extreme $\mu$ values, the attack methods fail to significantly reduce the accuracy. }
    \label{fig:lfr_combination}
\end{figure}

\begin{table*}[t]
\footnotesize
\centering
\caption{Micro F1-scores for node classification (three edge flip strategies: A: Addition, R: Removal, C: Combination) and average precision for link prediction (LP) on real world and  LFR networks ($\mu=0.3$). For link prediction, we only report results for edge deletion which has major impact on the embedding methods (effect of other edge flipping strategies is not shown here due to lack of space). The results of the best method and the second ranked  method are highlighted in bold and red, respectively. {\bf Lower value implies more efficiency to attack the embedding methods}.  \label{tab:node_classification_real}}
\scalebox{0.7}{
\begin{tabular}{@{}l|l|lll|l|lll|l|lll|l|lll|l|lll|l@{}}
\toprule
    & \multirow{2}{*}{Method}  & \multicolumn{4}{c}{SVD DeepWalk} & \multicolumn{4}{|c}{Skipgram DeepWalk} & \multicolumn{4}{|c}{Node2Vec} & \multicolumn{4}{|c}{LINE} & \multicolumn{4}{|c}{GCN} \\ \cline{3-22}
                              &                         & A                              & R                       & C                       & LP                     & A                       & R                       & C                       & LP                      & A                      & R                       & C                       & LP                      & A                      & R                              & C                      & LP                      & A                      & R                      & C                      & LP                      \\ 
\midrule
\multirow{5}{*}{\rot{(a) Cora}}     
& Clean                   & 0.82                           & 0.82                    & 0.82                    & 0.95                   & 0.81                    & 0.80                    & 0.81                    & 0.95                    & 0.77                   & 0.79                    & 0.76                    & 0.95                    & 0.76                   & 0.74                           & 0.73                   & 0.95                    & 0.79                   & 0.78                   & 0.79                   & 0.59                    \\
& Random                  & 0.76                           & 0.81                    & 0.78                    & \textcolor{red}{0.90}  & 0.74                    & 0.80                    & 0.76                    & \textcolor{red}{0.92 }  & 0.70                   & 0.77                    & 0.71                    & 0.91                    & \textcolor{red}{0.69}  & 0.71                           & 0.72                   & \textcolor{red}{0.91}   & 0.76                   & 0.78                   & 0.76                   & 0.53                    \\
& UNSUP                   & 0.81                           & \textbf{0.76}           & 0.76                    & \textbf{0.89}          & 0.79                    & \textbf{0.73}           & 0.72                    & 0.93                    & 0.75                   & \textbf{0.68}           & 0.67                    & \textcolor{red}{0.90}   & 0.73                   & \textbf{0.62}                  & \textbf{0.61}          & 0.93                    & 0.77                   & \textcolor{red}{0.75}  & 0.74                   & \textbf{0.46}           \\
& VIKING$^s$              & \textcolor{red}{0.73}          & 0.79                    & \textcolor{red}{0.74}   & \textbf{0.89}          & \textcolor{red}{0.70 }  & 0.77                    & \textcolor{red}{0.70}   & \textcolor{red}{0.92}   & \textbf{0.64}          & 0.74                    & \textcolor{red}{0.59 }  & \textbf{0.89}           & \textcolor{red}{0.69}  & 0.72                           & 0.69                   & \textcolor{red}{0.91}   & \textcolor{red}{0.74}  & 0.78                   & \textcolor{red}{0.73}  & 0.52                    \\
& VIKING                  & \textbf{0.72}                  & \textcolor{red}{0.77}   & \textbf{0.72}           & \textbf{0.89}          & \textbf{0.68}           & \textcolor{red}{0.75}   & \textbf{0.68}           & \textbf{0.91}           & \textcolor{red}{0.65}  & \textcolor{red}{0.71 }  & \textbf{0.58}           & \textcolor{red}{0.90}   & \textbf{0.67}          & \textcolor{red}{0.69 }         & \textcolor{red}{0.68}  & \textbf{0.90}           & \textbf{0.73}          & \textbf{0.74}          & \textbf{0.72}          & \textcolor{red}{0.49}   \\ 
\midrule
\multirow{5}{*}{\rot{(b) PolBlogs}} 
& Clean                   & 0.95                           & \textcolor{red}{0.95 }  & 0.95                    & 0.77                   & 0.95                    & \textcolor{red}{0.95}   & 0.95                    & 0.46                    & 0.95                   & 0.95                    & 0.95                    & 0.37                    & \textcolor{red}{0.93}  & 0.94                           & 0.93                   & 0.63                    & 0.96                   & 0.96                   & 0.96                   & \textbf{0.35}           \\
& Random                  & 0.94                           & \textcolor{red}{0.95}   & 0.95                    & \textbf{0.76}          & 0.93                    & \textcolor{red}{0.95 }  & 0.95                    & 0.46                    & 0.94                   & 0.95                    & \textcolor{red}{0.94}   & 0.41                    & \textcolor{red}{0.93}  & 0.94                           & 0.93                   & 0.64                    & 0.94                   & 0.95                   & 0.95                   & 0.39                    \\
& UNSUP                   & 0.95                           & \textcolor{red}{0.95}   & 0.94                    & 0.77                   & 0.95                    & \textcolor{red}{0.95 }  & 0.95                    & 0.46                    & 0.95                   & 0.94                    & 0.95                    & \textbf{0.32}           & \textcolor{red}{0.93}  & \textcolor{red}{0.91 }         & \textcolor{red}{0.91}  & 0.65                    & \textbf{0.80}          & \textcolor{red}{0.93}  & \textcolor{red}{0.94}  & 0.40                    \\
& VIKING$^s$              & \textcolor{red}{0.84}          & \textbf{0.90}           & \textcolor{red}{0.84 }  & \textbf{0.76}          & \textcolor{red}{0.83 }  & \textbf{0.89}           & \textcolor{red}{0.83 }  & \textbf{0.45}           & \textcolor{red}{0.85}  & \textcolor{red}{0.90}   & \textbf{0.84}           & 0.36                    & \textbf{0.90}          & 0.89                           & \textbf{0.90}          & \textbf{0.62}           & \textcolor{red}{0.90}  & \textbf{0.90}          & \textbf{0.90}          & \textcolor{red}{0.36}   \\
& VIKING                  & \textbf{0.83}                  & \textbf{0.90}           & \textbf{0.83}           & \textbf{0.76}          & \textbf{0.81}           & \textbf{0.89}           & \textbf{0.81}           & \textbf{0.45}           & \textbf{0.82}          & \textbf{0.89}           & \textbf{0.84}           & \textcolor{red}{0.33 }  & \textbf{0.90}          & \textbf{0.88}                  & \textbf{0.90}          & \textcolor{red}{0.63 }  & \textcolor{red}{0.90}  & \textbf{0.90}          & \textbf{0.90}          & \textcolor{red}{0.36}   \\ 
\midrule
\multirow{5}{*}{\rot{(c) CiteSeer}} 
& Clean                   & 0.69                           & 0.69                    & 0.69                    & 0.88                   & 0.67                    & 0.66                    & 0.66                    & 0.95                    & 0.66                   & 0.65                    & 0.66                    & 0.92                    & 0.58                   & 0.59                           & 0.57                   & 0.95                    & 0.64                   & 0.63                   & 0.64                   & 0.40                    \\
& Random                  & 0.56                           & 0.60                    & 0.55                    & \textbf{0.72}          & 0.53                    & 0.57                    & 0.54                    & \textcolor{red}{0.82}   & 0.51                   & 0.57                    & 0.52                    & 0.80                    & 0.51                   & 0.57                           & 0.52                   & \textcolor{red}{0.81}   & \textcolor{red}{0.55}  & 0.57                   & \textcolor{red}{0.55}  & \textcolor{red}{0.09}   \\
& UNSUP                   & 0.63                           & \textbf{0.52}           & \textcolor{red}{0.52 }  & 0.77                   & 0.55                    & \textbf{0.50}           & \textcolor{red}{0.48}   & 0.91                    & 0.53                   & \textbf{0.45}           & \textcolor{red}{0.47}   & 0.85                    & \textcolor{red}{0.49}  & \textbf{0.38}                  & \textbf{0.38}          & 0.91                    & 0.59                   & \textbf{0.52}          & 0.59                   & 0.10                    \\
& VIKING$^s$              & \textcolor{red}{0.54}         & \textcolor{red}{0.58}   & 0.54                    & 0.76                   & \textcolor{red}{0.51}   & 0.60                    & 0.51                    & 0.85                    & \textcolor{red}{0.49}  & 0.57                    & 0.49                    & \textcolor{red}{0.77}   & 0.50                   & \textcolor{red}{0.49}          & 0.51                   & 0.84                    & \textcolor{red}{0.55}  & 0.58                   & \textcolor{red}{0.55}  & 0.10                    \\
& VIKING                  & \textbf{0.50}                  & 0.60                    & \textbf{0.49}           & \textcolor{red}{0.74}  & \textbf{0.46}           & \textcolor{red}{0.56}   & \textbf{0.46}           & \textbf{0.81}           & \textbf{0.44}          & \textcolor{red}{0.56}   & \textbf{0.45}           & \textbf{0.76}           & \textbf{0.45}          & 0.50                           & \textcolor{red}{0.46}  & \textbf{0.79}           & \textbf{0.51}          & \textcolor{red}{0.55}  & \textbf{0.51}          & \textbf{0.08}           \\ 
\midrule
\multirow{5}{*}{\rot{(d) LFR}}     
& Clean                   & 0.63                           & 0.63                    & 0.63                    & 0.63                   & 0.75                    & 0.74                    & 0.75                    & \textbf{0.35}           & \textcolor{red}{0.59}  & 0.62                    & 0.60                    & \textcolor{red}{0.34}   & 0.71                   & 0.71                           & 0.71                   & 0.38                    & 0.59                   & 0.58                   & 0.59                   & 0.13                    \\
& Random                  & 0.62                           & 0.61                    & 0.62                    & 0.65                   & 0.73                    & 0.73                    & 0.72                    & 0.45                    & 0.60                   & 0.60                    & \textcolor{red}{0.57}   & \textcolor{red}{0.34}   & \textcolor{red}{0.70}  & 0.71                           & 0.70                   & 0.37                    & 0.60                   & \textbf{0.50}          & 0.58                   & 0.13                    \\
& UNSUP                   & 0.59                           & \textcolor{red}{0.58}   & \textcolor{red}{0.58}   & 0.62                   & 0.73                    & 0.70                    & \textcolor{red}{0.71}   & 0.48                    & \textcolor{red}{0.59}  & \textcolor{red}{0.59}   & \textbf{0.56}           & \textbf{0.32}           & \textbf{0.67}          & \textbf{0.66}                  & \textcolor{red}{0.67}  & 0.41                    & \textcolor{red}{0.52}  & \textbf{0.50}          & 0.60                   & 0.13                    \\
& VIKING$^s$              & \textcolor{red}{0.57}          & \textcolor{red}{0.58}   & \textcolor{red}{0.58}   & \textcolor{red}{0.60}  & \textcolor{red}{0.72}   & \textcolor{red}{0.68}   & \textcolor{red}{0.71}   & \textcolor{red}{0.36}   & \textcolor{red}{0.59}  & \textbf{0.58}           & 0.58                    & 0.35                    & \textbf{0.67}          & \textcolor{red}{0.69}          & \textcolor{red}{0.67}  & \textcolor{red}{0.33}   & 0.62                   & \textbf{0.57}          & \textcolor{red}{0.55}  & \textbf{0.10}           \\
& VIKING                  & \textbf{0.56}                  & \textbf{0.55}           & \textbf{0.55}           & \textbf{0.59}          & \textbf{0.68}           & \textbf{0.63}           & \textbf{0.67}           & 0.38                    & \textbf{0.57}          & \textcolor{red}{0.59}   & \textbf{0.56}           & \textcolor{red}{0.34}   & \textbf{0.67}          & \textbf{0.66}                  & \textbf{0.66}          & \textbf{0.32}           & \textbf{0.50}          & 0.60                   & \textbf{0.53}          & \textcolor{red}{0.11}   \\ 
\midrule
\multirow{5}{*}{\rot{(e) FFire}}    
& Clean                   & 0.76                           & 0.76                    & 0.76                    & 0.69                   & 0.67                    & 0.65                    & 0.67                    & 0.76                    & 0.51                   & 0.49                    & 0.49                    & 0.55                    & 0.67                   & 0.67                           & 0.66                   & 0.70                    & 0.82                   & 0.82                   & 0.82                   & 0.55                    \\
& Random                  & 0.57                           & 0.73                    & \textcolor{red}{0.60}   & 0.64                   & \textcolor{red}{0.53}   & 0.65                    & 0.55                    & 0.69                    & \textbf{0.33}          & \textcolor{red}{0.42}   & 0.40                    & 0.46                    & \textbf{0.54}          & 0.65                           & 0.58                   & 0.61                    & \textcolor{red}{0.58}  & \textcolor{red}{0.65}  & 0.60                   & 0.22                    \\
& UNSUP                   & 0.70                           & \textbf{0.63}           & 0.64                    & 0.60                   & 0.61                    & \textbf{0.56}           & 0.57                    & 0.73                    & 0.43                   & \textcolor{red}{0.42}   & 0.41                    & 0.49                    & 0.63                   & \textbf{0.50}                  & \textbf{0.50}          & 0.58                    & 0.69                   & 0.68                   & 0.66                   & \textcolor{red}{0.17}   \\
& VIKING$^s$              & \textbf{0.56}                  & \textcolor{red}{0.67}   & \textbf{0.57}           & \textbf{0.53}          & \textbf{0.52}           & \textcolor{red}{0.58}   & \textbf{0.52}           & \textbf{0.62}           & \textcolor{red}{0.37}  & \textbf{0.39}           & \textbf{0.29}           & \textbf{0.41}           & \textcolor{red}{0.55}  & 0.55                           & \textcolor{red}{0.54}  & \textbf{0.52}           & \textcolor{red}{0.58}  & \textbf{0.59}          & \textcolor{red}{0.59}  & \textbf{0.13}           \\
& VIKING                  & \textcolor{red}{0.57}          & \textbf{0.63}           & \textbf{0.57}           & \textcolor{red}{0.57}  & \textcolor{red}{0.53}   & \textcolor{red}{0.58}   & \textcolor{red}{0.54}   & \textcolor{red}{0.66}   & 0.38                   & \textcolor{red}{0.42}   & \textcolor{red}{0.37}   & \textcolor{red}{0.45}   & 0.57                   & \textcolor{red}{0.53}          & 0.56                   & \textcolor{red}{0.54}   & \textbf{0.57}          & \textbf{0.59}          & \textbf{0.58}          & \textbf{0.13}           \\
\bottomrule
\end{tabular}}
\end{table*}

\noindent\textbf{\underline{Performance over tasks}:}
Table \ref{tab:node_classification_real} shows that VIKING performs well over both node classification and link prediction. However, even its semi-supervised setting, VIKING$^s$  performs comparable to the supervised counterpart.
On real-world datasets, \name\ performs extremely well, achieving at least 13\% score reduction on each dataset using Skipgram DeepWalk. The attack strategy is also effective across various random walk based methods such as Node2Vec (LINE), resulting in decrements of up to $18\%$ ($9\%$) in Cora, $13\%$ ($6\%$) in PolBlogs, and $18\%$ ($20\%$) in Citeseer network. In most tasks across all the networks, \name\ and \name$^s$ outperform the baseline attacking strategies.

\begin{figure}[!t]
\centering
  \includegraphics[width=0.8\textwidth]{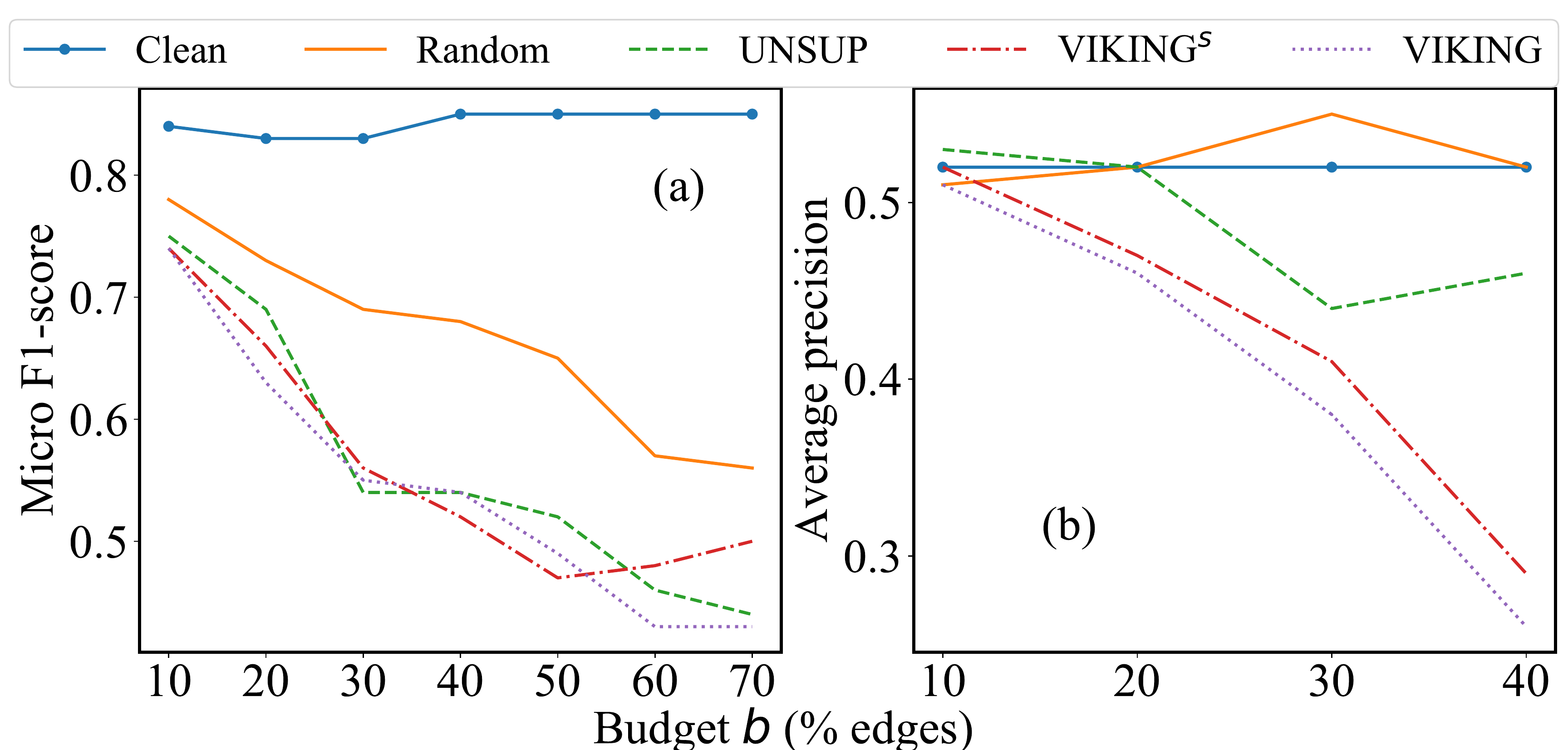}
  \caption{Performance variation with budget for (a) node classification, and (b) link prediction on PolBlogs network. For both tasks, the performance shows a sharp decrease with VIKING outperforming others.}
  \label{fig:varying_edges}
\end{figure}

Since the absolute budget for all the networks is same \textcolor{black}{(except Forest Fire in which the number of edges is already quite low)}, i.e., 1000 (Table \ref{tab:datasets}), we see that the least performance degradation is in PolBlogs, which is expected because the fraction of edges flipped is small. This is also confirmed by Figure \ref{fig:varying_edges}. Also, the number of unique labels does not seem to affect the performance of our attack. \name\ is successful in all three networks where the number of unique labels ranges from 2 to 7. 

\noindent\textbf{\underline{Effect of community mixing ($\mu$) in LFR networks)}:}
Figure \ref{fig:lfr_combination} shows the performance on the LFR network by varying $\mu$ from $0.1$ to $0.7$. Similar pattern is observed with \name\ outperforming others at every value of $\mu$. We report values on LFR graphs at a non-extreme generator parameter $\mu=0.3$ in Table \ref{tab:node_classification_real}. Even for synthetic network, \name\ and \name$^s$ dominate as the strategies for attacking with a decrease in performance of $4\%$ and $5\%$ on Node2Vec and LINE, respectively. \name\ is clearly \textcolor{black}{a} better attacking strategy across the various embedding methods considered. 

\noindent\textbf{\underline{Side-by-side diagnostics}:}
Interestingly, for link prediction, sometimes the results seem counter intuitive. For example, in Skipgram DeepWalk embeddings for LFR graphs, we observe that on all the poisoned graphs, the performance has actually improved (across all poisoning strategies) by a score of $1-13\%$. However, the increase is least in case of \name\ and \name$^s$. This is possibly a result of the sparsity of the graph. All attacking strategies flip edges in order to reduce density; however the budget is not enough and the resultant poisoned graphs form distinctly differentiable structures (e.g., triad to path), resulting in a net increase of scores. This remains a limitation of \name\ along with other strategies. This does not affect the node classification task though for most embedding methods.

\noindent\textbf{\underline{Effect of varying budget $b$}:}
Since PolBlogs has the maximum number of edges amongst all the networks used, we use this network to observe how the number of flips affects the accuracy of the downstream task. Candidate edges have both addition and removal involved. Figure \ref{fig:varying_edges} clearly shows that \name\ is better than other alternatives \textcolor{black}{in both node classification (Figure \ref{fig:varying_edges}(a)) and link prediction (Figure \ref{fig:varying_edges}(b)).}

\noindent\textbf{\underline{Inspecting performance of semi-supervised approach}:}
So far, we assumed full knowledge of the underlying network (training dataset only). Here, we discuss the observations without using full knowledge of node labels. Table \ref{tab:node_classification_real} also shows the performance of various embedding methods in the semi-supervised attack setting with VIKING$^s$. We observe that the attack is successful even with partial knowledge. An attacker does not need to know the labels of all the nodes; information of even 10\% labels can be effective. We generate embeddings using an unsupervised embedding method (DeepWalk). We then use these 10\% labels to predict the other 90\% labels. The detailed attacking strategy is discussed in Section \ref{vikingminus}.

\begin{figure}[!t]
    \centering
    \includegraphics[width=0.8\textwidth]{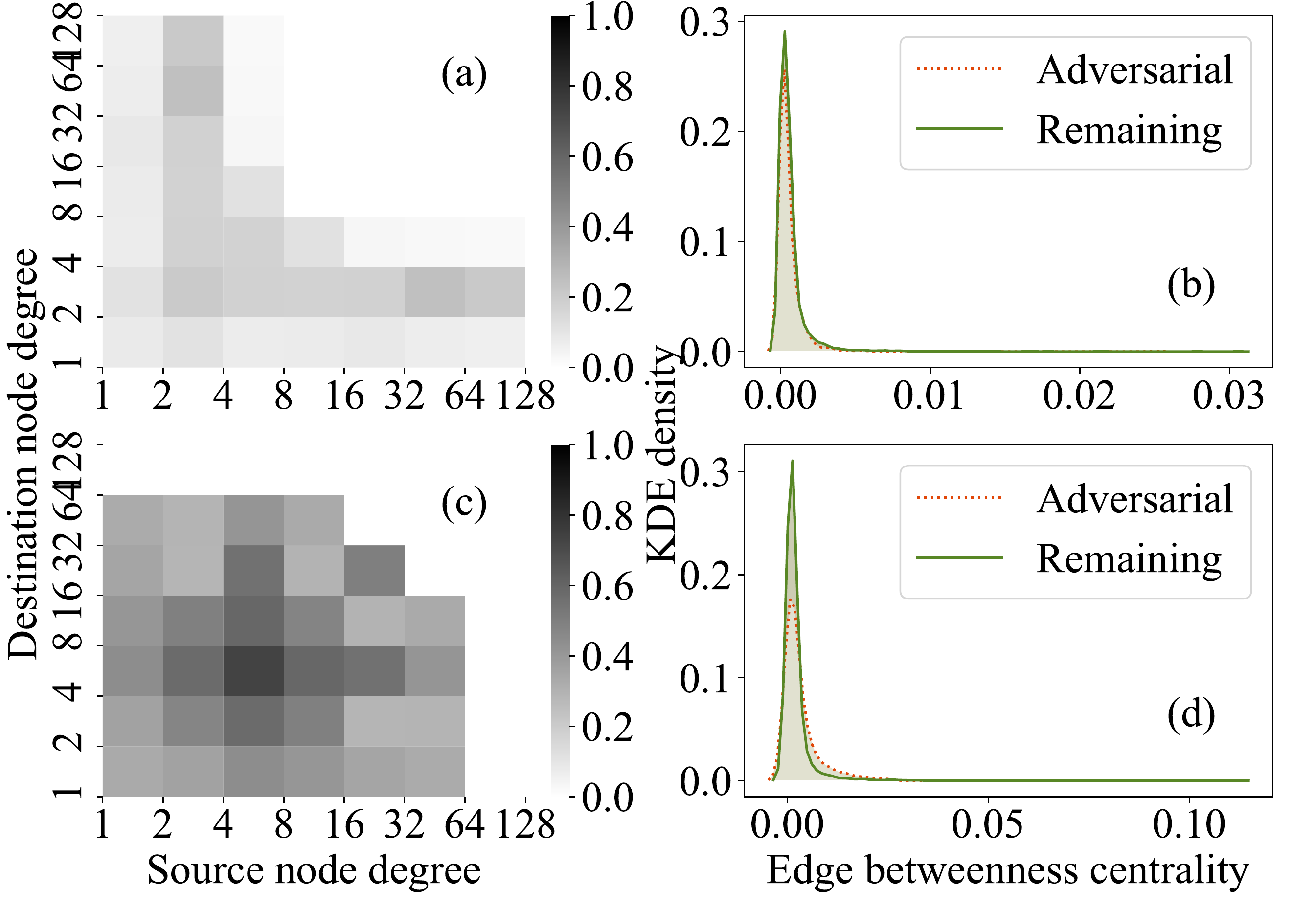}
    \caption{Analysis of adversarial edge properties: (a, c) degree distribution of nodes on adversarial edges;  (b, d) betweenness centrality distribution of adversarial edges. The distribution of properties such as node degrees and edge centralities do not suggest any intelligent heuristic. (a)-(b) correspond to Cora; (c)-(d) correspond to Citeseer.}
    \label{fig:edge_analysis}
\end{figure}

\noindent\textbf{\underline{Analysing adversarial edges}:}
To investigate if the edges selected by \name \ have any distinct characteristics that may be used to eliminate usage of labels, we analyse  node degrees and edge betweenness centrality values of the selected edges from Cora. Figures \ref{fig:edge_analysis} (a, c) show a logarithmically binned heatmap of fraction of adversarial edges w.r.t. total number of edges for each degree. Nodes across all -- low, medium and high -- degrees are used in the adversarial edges. Figures \ref{fig:edge_analysis} (b, d) show the distribution of edge centrality of both adversarial and non-adversarial edges. The distribution of both adversarial (red) and non-adversarial (green) edges is extremely close. Both the distributions peak at the same edge betweenness centrality value ($0.0005$ and $0.001$ after normalisation in Figures \ref{fig:edge_analysis}(b) and \ref{fig:edge_analysis}(d), respectively) and have a similar spread of distribution. From these plots, we can conclude that selected edges affect nodes with different degrees, and adversarial edges have an extremely similar distribution as compared to the rest of the edges. This analysis shows that developing attack heuristics is not trivial and simple network measures (such as degree, centrality) are not sufficient to characterize the edges selected by \name.

\section{Conclusions\label{conclusion}}
We studied the robustness of random-walk based embeddings and measured the performance of our heuristic method for supervised poisoning attacks on network data. We presented \name, a generic framework for adversarial poisoning attacks on networks. The experiments performed on multiple datasets included comparisons with existing methods to establish the effectiveness of \name. Furthermore, a semi-supervised extension of \name\ was also tested which demonstrates the efficacy of poisoning attacks even with partial label knowledge. Based on the current study, we conclude that in network science an important need is to develop robust embedding methods for large-scale networks.

{\small
\bibliographystyle{splncs04}
\bibliography{viking}}

\end{document}